\documentclass{article}

\usepackage[round]{natbib}

\usepackage{mathtools}
\usepackage{amsfonts}       
\usepackage{amsthm}
\usepackage{amsmath}
\usepackage{amssymb}
\usepackage{bbm}
\usepackage{bm}
\newtheorem{thm}{Theorem}
\newtheorem{lem}{Lemma}
\newtheorem{prop}{Proposition}

\newtheorem{claim}{Claim}
\newtheorem{aspt}{Assumption}

\theoremstyle{definition}

\DeclareMathOperator*{\argmax}{arg\,max}
\DeclareMathOperator*{\argmin}{arg\,min}
\newcommand\numberthis{\addtocounter{equation}{1}\tag{\theequation}}

\newcommand{\E}{\mathbb{E}}

\usepackage{algorithm}
\usepackage{algcompatible}
\algnewcommand\algorithmicreturn{\textbf{return}}
\algnewcommand\RETURN{\State \algorithmicreturn}%

\usepackage{xcolor}
\newcount\comments  
\newcommand{\genComment}[2]{\ifnum\comments=1{\textcolor{#1}{\textsf{\footnotesize #2}}}\fi}

\newcommand{\ziping}[1]{\genComment{blue}{[ZX:#1]}}

\begin{document}

%

%

\title{On the Statistical Benefits of Curriculum Learning}

\author{Ziping Xu, Ambuj Tewari\\University of Michigan, Department of Statistics}
\maketitle


\begin{abstract}
    Curriculum learning (CL) is a commonly used machine learning training strategy. However, we still lack a clear theoretical understanding of CL's benefits.
    In this paper, we study the benefits of CL in the multitask linear regression problem under both structured and unstructured settings. For both settings, we derive the minimax rates for CL with the oracle that provides the optimal curriculum and without the oracle, where the agent has to adaptively learn a good curriculum. 
    Our results reveal that adaptive learning can be fundamentally harder than the oracle learning in the unstructured setting, but it merely introduces a small extra term in the structured setting. To connect theory with practice, we provide justification for a popular empirical method that selects tasks with highest local prediction gain by comparing its guarantees with the minimax rates mentioned above.
\end{abstract}

\section{Introduction}
It has long been realized that we can design more efficient learning algorithms if we can make them learn on multiple tasks. Transfer learning, multitask learning and meta-learning are just few of the sub-areas of machine learning where this idea has been pursued vigorously. Often the goal is to minimize the weighted average losses over a set of tasks that are expected to be similar. While previous literature often assumes a predetermined (and often equal) number of observations for all the tasks, in many applications, we are allowed to decide the \textit{order} in which the tasks are presented and the \textit{number of observations} from each task. Any strategy that tries to improve the performance with a better task scheduling is usually referred to \textbf{curriculum learning (CL)} \citep{bengio2009curriculum}. The agent that schedules tasks at each step is often referred as the \textit{task scheduler}.

Though curriculum learning has been extensively used in modern machine learning \citep{gong2016multi,sachan2016easy,tang2018attention,narvekar2020curriculum}, there is very little theoretical understanding of the actual benefits of CL. We also do not know whether the heuristic methods used in many empirical studies can be theoretically justified. Even the problem itself has not been rigorously formulated. To address these challenges, we first formulate the curriculum learning problem in the context of the linear regression problem. We analyze the minimax optimal rate of CL in two settings: an unstructured setting where parameters of different tasks are arbitrary and a structured setting where they have a low-rank structure. Finally we discuss the theoretical justification of a popular heuristic task scheduler that greedily selects tasks with highest local prediction gain.

\section{Background}
We would like to point out previous work on three crucial aspects of CL: two types of benefits one may expect from CL, task similarities assumptions, and task scheduler used in empirical studies. 

\begin{figure}
    \centering
    \includegraphics[width = 0.4\textwidth]{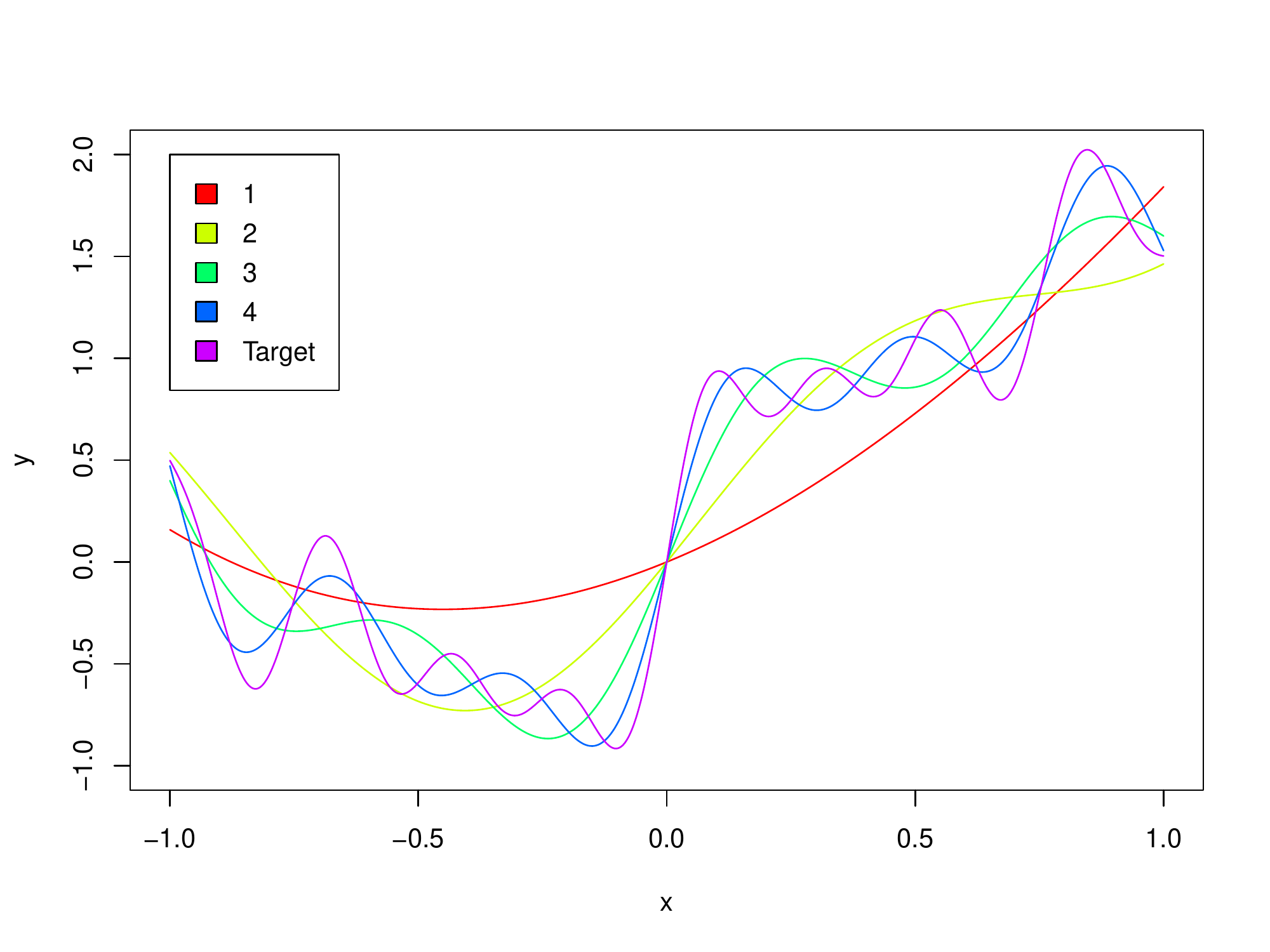}
    \caption{An example of tasks with increasing non-convexity. Solid lines of different colors represent the true object functions of different tasks.}
    \label{fig:1}
\end{figure}

\paragraph{Two types of benefits.} There are two distinct ways to understand the benefits of CL. From the perspective of optimization, some papers argue that the benefits of curriculum can be interpreted as learning from more convex and more smooth objective functions, which serves as a better initialization point for the non-convex target objective function \citep{bengio2009curriculum}. The order of task scheduling is essential here. As an example, Figure \ref{fig:1} shows the objective functions of a problem with four source tasks and one target task with increasing difficulty (non-convexity). Directly minimizing the target task (marked in purple line) using gradient descent can be hard due to the non-convexity. However, the simple gradient descent algorithm can converge to the global optima of the current task if it starts from the global optima of the previous task. We refer to the benefit that involves a faster convergence in optimization as \textit{optimization benefit}. Optimization benefits highly depend on the order of scheduling. Generally speaking, if one directly considers the empirical risk minimizer (ERM) which requires global minimization of empirical risk, there may not be any optimization benefit. 

The second type of benefit concerns the benefit brought by carefully choosing the number of observations from each task while independent of the order, which we call \textit{statistical benefit}. For example, we have two linear regression problems that are identical except for the standard deviation of the Gaussian noise on response variables. 
If we consider the OLS estimator on the joint dataset of the two tasks, there is a reduction in noise level when more samples are allocated to the task with a lower level, and the benefit is independent of the order by the nature of OLS. A \textit{statistical benefit} can be seen as any benefit one can get except for the reduction in the difficulties of optimization. 
\cite{weinshall2020theory} focused on a special curriculum learning task where each sample is considered a task and they analyzed the convergence rate on the samples of different noise levels. They analyzed the benefits on the convergence rate, which should count as the \textit{statistical benefits}.

In general, the two types of benefits can coexist. A good curriculum should account for both the non-convexity and the noise levels. However, due to the significantly different underlying mechanism in the two learning benefits. it is natural to study them separately. This paper will focus on the analysis of the \textbf{statistical benefits}. Thus, we analyze algorithms that map datasets to an estimator for each task that may involve finding global minima of the empirical errors for non-convex functions. 

\paragraph{Similarity assumptions.} We discussed the problem with two almost identical tasks, where we can achieve perfect transfer and the trivial curriculum that allocates all the samples to the simpler task is optimal. However, tasks are generally not identical. Understanding how much benefits the target task can gain from learning source tasks has been a central problem in transfer learning and multitask learning literature. The key is to propose meaningful similarity assumptions. 

Let $\mathcal{X}$ and $\mathcal{Y}$ be the input and output space. Assume we have $T$ tasks with data distributions $D_1 \dots, D_T$ over $\mathcal{X} \times \mathcal{Y}$. Let $(X_t, Y_t)$ be a sample from task $t$. 
Let $f_t = \mathbb{E}[Y_t \mid X_t]$, a mapping $\mathcal{X} \mapsto \mathcal{Y}$, be the mean function. In this paper, we adopt the simple parametric model on the mean function with $f_t$ represented by parameter $\theta_t^* \in \mathbb{R}^{d}$. 

We consider two scenarios: structured and unstructured. In Section \ref{sec:unstructured}, we adopt simple linear regression models and do not assume any further internal structure on the true parameters. Two tasks are similar if $\|\theta_{t_1}^* - \theta_{t_2}^*\|_2$ is small. A learned parameter is directly transferred to the target task. This setting has been applied in many previous studies \citep{yao2018direct,bengio2009curriculum,xu2021decision}.
In Section \ref{sec:MRL}, we study the multitask representation learning setting \citep{maurer2016benefit,tripuraneni2020theory,xu2021representation}, where a stronger internal structure is assumed. To be specific, we write $f_t(x) = x^T B^* \beta_t^*$, where $x\in\mathbb{R}^{d}$, $B^* \in \mathbb{R}^{d \times k}$ is the linear representation mapping and $\beta_t^* \in \mathbb{R}^{k}$ is the task specific parameter. Generally, the input dimension is much larger than the representation dimension ($d \gg k$).

These two settings, while representative, do not exhaust all of the settings in the literature. We refer the reader to \cite{teshima2020few} for a brief summary of theoretical assumptions on the task similarity. 

\paragraph{Task schedulers.} 


Many empirical methods have been developed to automatically schedule tasks. \cite{liu2020adaptive} designed various heuristic strategies for task selection for computer vision tasks. \cite{cioba2021distribute} discussed several meta-learning scenarios where the optimal data allocations are different, which interestingly aligns with our theoretical results. For a more general use, one major family of task scheduler is based on the intuition that the task scheduler should select the task that leads to the highest local gain on the target loss \citep{graves2017automated}. Since the accurate prediction gain is not accessible, online decision-making algorithms (bandit and reinforcement learning) are frequently used to adaptively allocate samples \citep{narvekar2020curriculum}. However, there is no theoretical guarantee that such greedy algorithms can lead to the optimal curriculum.

\paragraph{Notation.} For any positive integer $n$, we let $[n] = \{1, \dots, n\}$. We use the standard $O(\cdot), \Omega(\cdot)$ and $\Theta(\cdot)$ notation to hide universal constant factors. We also use $a \lesssim b$ and $a \gtrsim b$ to indicate $a = O(b)$ and $b = \Omega(b)$.



\section{Unstructured Linear Regression}
\label{sec:unstructured}
In this section, we study the problem of learning from $T$ tasks to generate an estimate for a single target task. 
\subsection{Formulations}
\label{sec:setup}
We consider linear regression tasks. Let $1 \dots, T$ denote $T$ tasks. Let $\theta_t^* \in \mathbb{R}^d$ denote the true parameter of task $t$. The response $Y_t$ of task $t$ is generated in the following manner
$$
    Y_t = X_t^T \theta_t^* + \epsilon_t,
$$
where $\epsilon_t$ is assumed to be the Gaussian noise with $\mathbb{E}[\epsilon_t^2] = \sigma_t^2$ and 
$
    X_t \sim \mathcal{N}(\bm{0}, \Sigma_t),
$
where $\Sigma_t \in \mathbb{R}^{d \times d}$ is the covariance matrix that is positive definite. 
Any task, therefore, can be fully represented by a triple $(\theta_t^*, \sigma_t, \Sigma_t)$. 

Throughout the paper, we are more interested in the unknown parameters rather than the covariate distribution or the noise level.
We simply denote $\bm{\theta}^* \in \mathbb{R}^{d \times T}$ the parameters of a \textit{problem} ($T$ tasks) and let $\bm{\theta}_t^*$ be the $t$-th column of the matrix. 

We make a uniform assumption on the covariance matrix of input variables. The same assumption is also used by \cite{du2020few}.
\begin{aspt}[Coverage of covariate distribution]
\label{aspt:coverage}
We assume that all $C_0 I_d\succ \Sigma_t \succ C_1I_d$ 
for some constant $C_0, C_1 > 0$ and any $t\in [T]$.
\end{aspt}

\paragraph{Goal.} Let $S_t^{n}$ be random $n$ samples from task $t$.  Let $l: \mathbb{R} \times \mathbb{R} \mapsto \mathbb{R}$ be a loss function and
$L_t(\theta) = \mathbb{E}[l(X^T \theta, Y)]$ be the expected loss of a given hypothesis $\theta$ evaluated on task $t$. Moreover, we denote the excess risk by
$$
    G_t(\theta) = L_t(\theta) - \inf_{\theta'}L_{t}(\theta').
$$ 
Our goal in this section is to minimize the expected loss of the last task $T$, which we call the \textbf{target task}. Throughout the paper, we use square loss function.

\paragraph{Transfer distance.} Algorithms tend to perform better when the tasks are similar to each other, such that any observations collected from non-target task bear less transfer bias. We define \textit{transfer distance} between tasks $t_1, t_2$ as $\Delta_{t_1, t_2} = \|\theta_{t_1}^* - 
\theta_{t_2}^*\|_2$. 

It is not fair to compare the performances between problems with different transfer distances. To study a minimax rate, we are interested in the worst performance over a set of problems with similar transfer distance. Let $\bm{Q} \in \mathbb{R}^{T}$ be the distance vector encoding the upper bounds on the distance between the target task to any task. 
We define the hypothesis set with known transfer distance as $\Theta(\bm{Q}) =  \cup_{p}\{\bm{\theta}^*: \|\bm{\theta}^*_t - \bm{\theta}_T^*\|_2 \leq \bm{Q}_{t}\} \subset \mathbb{R}^{T \times d}$.
The hypothesis set with unknown transfer distance can be defined as $\tilde\Theta(\bm{Q}) =  \cup_{p}\{\bm{\theta}^*: \|\bm{\theta}_{p_t}^* - \bm{\theta}_T^*\|_2 \leq \bm{Q}_{p_t}\}$, where $p \in [T]^T$ is any permutation of $[T]$. We say this hypothesis set has unknown transfer distance because even if there exists some small $\bm{Q}_t$ such that the transfer distance is low, an agent does not know which task has the low transfer distance.

\paragraph{Curriculum learning and task scheduler.} This paper concerns only the statistical learning benefits. Since the order of selecting tasks does not affect the outcome of the algorithm, we denote a curriculum by $\bm{c} \in [N]^T$, where each $\bm{c}_t$ is the total number of observations from task $t$ and $\sum_t \bm{c}_t = N$. Note that $\bm{c}$ can consist of random variables depending on the task scheduler. The set of all the curriculum with a total number of observations $N$ is denoted by $\mathcal{C}_N = \{\bm{c} \in [N]^T: \sum_t{\bm{c}_t} = N\}$. 

Any curriculum learning involves a multitask learning algorithm, which is defined as a mapping $\mathcal{A}$ from a set of datasets $(S_{1}^{n_1}, \dots, S_{T}^{n_T})$ to a hypothesis $\theta$ for the target task. 

A task scheduler runs the following procedure. At the start of the step $i \in [N]$, we have $n_{i, 1}, \dots, n_{i, T}$ observations from each task. The task scheduler $\mathcal{T}$ at step $i$ is defined as a mapping from the past observations $(S_1^{n_{i, 1}}, \dots, S_T^{n_{i, T}})$ to a task index. Then a new observation from the selected task $\mathcal{T}(S_1^{n_{i, 1}}, \dots, S_T^{n_{i, T}}) \in [T]$ is sampled.

\paragraph{Minimax optimality and adaptivity.} One of the goals of this work is to understand the minimax rate of the excess risk on the taregt task over all the possible combinations of multitask learning algorithms and task schedulers. We first attempt to understand a limit of that rate by considering an oracle scenario that provides the optimal curriculum for any problem.

Rigorously, we denote the loss of a fixed curriculum $\bm{c} \in \mathcal{C}_N$ with respect to a fixed algorithm $\mathcal{A}$ and problem $\bm{\theta}$ by
$$
    R_T^N(\bm{c}, \mathcal{A} \mid \bm{\theta}) = \mathbb{E}_{S_1^{\bm{c}_1}, \dots ,S_T^{\bm{c}_T}}G_T(\mathcal{A}(S_1^{\bm{c}_1}, \dots, S_T^{\bm{c}_T})).
$$

We define the following oracle rate, which takes infimum over all the possible fixed curriculum designs given a fixed task set with different $\bm{\theta}$ in a hypothesis set $\Theta$. 
\begin{align*}
    R_T^N(\Theta) \coloneqq
    \inf_{\mathcal{A}}\sup_{\bm{\theta} \in \Theta} \inf_{\bm{c} \in \mathcal{C}_N} R_T^N(c, \mathcal{A} \mid \bm{\theta}). \numberthis \label{equ:oracle}
\end{align*}

In general, the above oracle rate considers an ideal case, because the optimal curriculum depends on the unknown problem and any learning algorithm has to adaptively learn the problem to decide the optimal curriculum.

We ask the following question: can adaptively learned curriculum perform as well as the optimal one as in Equ. (\ref{equ:oracle})? To answer the question, we define the minimax rate for adaptive learning:
\begin{align*}
    \tilde{R}_T^N(\Theta) \coloneqq
    \inf_{\mathcal{A} } \inf_{\mathcal{T}} \sup_{\bm{\theta} \in \Theta} \mathbb{E} G_T(\mathcal{A}(S_1^{\bm{c}_{\mathcal{T}, 1}}, \dots, S_T^{\bm{c}_{\mathcal{T}, T}})),
    \numberthis \label{equ:adaptive}
\end{align*}
where $\bm{c}_{\mathcal{T}} \in \mathcal{C}_N$ is the curriculum adaptively selected by the task scheduler $\mathcal{T}$ and the expectation is taken over both datasets $S_1, \dots, S_T$ and $\bm{c}_{\mathcal{T}}$. 

In this section, we are interested in the oracle rate in ($\ref{equ:oracle}$) compared to some naive strategy that allocates all the samples to one task. This answers how much benefits one can achieve compared to some naive learning schedules. We are also interested the gap between Equ. (\ref{equ:oracle}) and Equ. (\ref{equ:adaptive}).

\subsection{Oracle rate}
In this section, we analyze the oracle rate defined in Equ. (\ref{equ:oracle}). We first give an overview of our results.
For any problem instance, there exists a single task $t$ such that the naive curriculum with $\bm{c}_t = N$ matches a lower bound for the oracle rate defined in Equ. (\ref{equ:oracle}).

For any task $t \in [T]$, its direct transfer performance of the OLS estimator on the target task $T$ can be roughly bounded by $\Delta_{t, T}^2 + {d\sigma_t^2}/N$.

Thus, our result implies that essentially, the goal of curriculum learning is to identify the best task that balance the transfer distance and the noise level.

\begin{thm}
\label{thm:oracle_lower}
Let $\bm{Q}$ be a fixed distance vector defined above. 
The oracle rate within $\Theta(\bm{Q})$ in Equ. (\ref{equ:oracle}) can be lower bounded by
\begin{equation}
    R_T^N({\Theta}(\bm{Q})) \gtrsim C_0 \min_{t}\{\bm{Q}_{t}^2 + \frac{d{\sigma}_t^2}{N}\}. \label{equ:thm1}
\end{equation}
\end{thm}

\paragraph{Proof highlights.} \cite{kalan2020minimax} showed a minimax rate of the transfer learning problem with only one source task. They considered three scenarios, which can be uniformly lower bounded by the right hand side of Equ. (\ref{equ:thm1}). Our analysis can be seen as an extension of their results to multiple source tasks. 
In general, let the rate in (\ref{equ:thm1}) be $\delta$ and $C > 0$ be a constant. Let $t^*$ be the best task indicated by (\ref{equ:thm1}). Any task $t$ with a large distance ($\bm{Q}_t > C\delta$) is not helpful to learn the target task. Thus, samples from these tasks can be discarded without reducing the performance. For any task $t$ with $\bm{Q}_t \leq C\delta$, we will show that any sample from task $t$ gives almost as much as information as the best task $t^*$ gives. Thus, one can replace them with a random sample from the best task $t^*$ without reducing the loss. Then the problem can be converted to a single source task problem, from where we follow the lower bound construction in \cite{kalan2020minimax}.

\subsection{Minimax rate for adaptive learning}


The problem can be hard when the transfer distance is unknown. We introduce an intuitive example to help understand our theoretical result. Assume we have three tasks including one target task and two source tasks. One of the two source tasks is identical to the target task. We have $n$ samples for both source tasks, while no observations from the target task. In this example, even if one of the source tasks is identical to the target task, no algorithm can decide which source task should be adopted, since we have no information from the target task. In other words, any algorithm can be as bad as the worst out of the two source tasks. This is not an issue when the transfer distance is known to the agent in the oracle scenario. This example implies that to adaptively gain information from source tasks, we will need sufficient information from the target task. Otherwise, there is risk of including information from tasks that contaminates the target task. Similarly, \cite{david2010impossibility} also showed that without any observations from the target task, domain adaptation is impossible.

More generally, even if we have some data from the target task, we will show that one is not able to avoid $\sigma_T^2$ term, the learning difficulty of the target task. Now we formally introduce our results. 
\begin{thm}
\label{thm:adp_lower}
Assume $T \geq 4$. Let $Q_{sub} > 0$. Let $\bm{Q}$ be a fixed distance vector that satisfies $\bm{Q}_{1} = 0$ and $\bm{Q}_{t} = Q_{sub} > 0$ for all $t = 2, \dots, T-1$. 
The minimax rate in Equ. (\ref{equ:adaptive}) can be lower bounded by
\begin{align*}
    \tilde{R}_T^N(\tilde{\Theta}(\bm{Q})) \gtrsim
      \min\{\frac{{\sigma}_T^2 \log(T)}{N}, Q^2_{sub}\} + \min_t\frac{d{\sigma}_t^2}{N}. 
    \numberthis \label{equ:thm2}
\end{align*}
\end{thm}

Theorem \ref{thm:adp_lower} implies that without knowing the transfer distance, any adaptively learned curriculum of any multitask learning algorithm will suffer an unavoidable loss of ${{\sigma}_T^2 \log(T)}/{N}$, when $Q_{sub}$ is large. Compared to the rate ${{\sigma}_T^2 d}/{N}$ without transfer learning, there is still a potential improvement of a factor of $d/\log(T)$ when $Q_{sub}$ and ${\sigma}_t^2$ are small.

\paragraph{Upper bound.} As we showed above, there is a potential improvement of $d/\log(T)$. This is because given the prior information that one of the source tasks is identical to the target task, the problem reduces from estimating a $d$-dimensional vector to identifying the best task from a candidate set, whose complexity reduces to $\log(T)$. 

In fact, a simple fixed curriculum could achieve the above minimax rate. Assume that any $\|\theta_t^*\|_2\leq C_2$ for some constant $C_2 > 0$. Let $\bm{c}_T = N/2$ and for all the other tasks $\bm{c}_t = N / (2T - 2)$. For each $t = 1, \dots T-1$, let $\tilde \theta_t$ be the OLS estimator using only its own samples. Let $\hat \theta_t$ be the projection of $\tilde \theta_t$ onto $\{\theta: \|\theta\|_2 \leq C_2\}$.
Then we choose one estimator from $t = 1, \dots, T-1$, that minimizes the empirical loss for the target task:
\begin{equation}
    t^* = \argmin_{t \in [T-1]} \sum_{i = 1}^{N/2} (Y_{T, i} - X_{T, i}^T 
    \hat\theta_{t})^2.
    \label{equ:guess}
\end{equation}

\begin{thm}
\label{thm:adp_upp}
Assume there exists a task $t$ such that $\Delta_{t, T} = 0$ and $\|\theta_t^*\|_2 \leq C_2$. With a probability at least $1-\delta$, $\hat \theta_{t^*}$ satisfies 
\begin{equation}
    G_T(\hat \theta_{t^*}) \lesssim C_0\log(T/\delta)\left(\frac{C_2\sigma_T^2 }{N} + \frac{dT\sigma_{t^*}^2}{C_1N} \right). \label{equ:thm3}
\end{equation}
\end{thm}

Note that $t^*$ is a random value. However, when all $t$ satisfy ${dT\sigma_t^2} < \Delta_{max}\sigma_T^2\log(T)$, the first term is the dominant term and our bound matches the lower bound in (\ref{equ:thm2}). This could happen when $\sigma_T^2 \gg \sigma_t^2$ for all $t = 1, \dots, T-1$. For a fixed problem instance, as long as $N$ is sufficiently large, one should be able to identify the optimal source task, which removes the dependence of $d$ in the second term above. To this end, we introduce another task scheduler based on task elimination in Appendix \ref{app:thm3}. 

\paragraph{General function class.}
As we mentioned before, though it is difficult to identify the good source tasks, the complexity of doing so is still lower than learning the parameters directly. We remark that this result can be generalized to any function class beyond linear functions. Keeping all the other setup unchanged, we assume that the mean function $f_t^* \in \mathcal{F}_t: \mathcal{X} \mapsto \mathcal{Y}$ for some input space $\mathcal{X}$ and output space $\mathcal{Y}$ shared by all the tasks. For convenience, we assume there is no covariate shift, i.e. the input distributions are the same. We give an analogy of Theorem \ref{thm:adp_upp}. 

\begin{aspt}[Assumption B in \cite{jin2021nonconvex}]
\label{aspt:gen}
Assume $l(\cdot, y)$ is $L_2$-strongly convex and $L_1$-Lipschitz at any $y \in \mathcal{Y}$. Furthermore, for all $x \in \mathbb{R}^d$ and $t \in [T]$,
$$
    \mathbb{E}[\nabla l(f_t^*(X), Y) \mid X = x] = 0.
$$
\end{aspt}

Assume we have $N / (2T - 2)$ observations for all tasks $t = 1, \dots, T$ and $N / 2$ observations for the target task. Let $\hat f_t$ be the empirical risk minimizer of the task $t$. Similarly to (\ref{equ:guess}), let 
$$
    t^* = \argmin_{t \in [T-1]} l^{N}_T(\hat f_t),
$$
where $l^{N}_T$ is the empirical loss on task $T$.
Let $L^* = \min_{t \in [T-1]} L_T(f_t^*)$ and $t' = \argmin_{t \in [T-1]} L_T(f_t^*)$. We will use Rademacher complexity to measure the hardness of learning a function class. We refer readers to \cite{bartlett2002rademacher} for the detailed definition of Rademacher complexity.

\begin{prop}
\label{prop:1}
Given the above setting and Assumption \ref{aspt:gen}, we have with a probability at least $1-\delta$,
\begin{align*}
    G_T(\hat f_{t^*}) \lesssim L^* + {\frac{L_1}{L_2}}\left(\mathcal{R}_{N/T}(\mathcal{F}_{t'}) + \sqrt{\frac{ \log(1/\delta)}{N/T}} \right),
\end{align*}
where $\mathcal{R}_{N}(\mathcal{F})$ is the Rademacher complexity of function space $\mathcal{F}$.
\end{prop}
This bound improves the bound for single target task learning, which scales with $\mathcal{R}_{N}(\mathcal{F}_T)$, when $\mathcal{R}_{N}(\mathcal{F}_T) \gg \mathcal{R}_{N/T}(\mathcal{F}_{t^*})$. The underlying proof idea is still that identifying good tasks is easier than learning the model itself.

\section{Structured Linear Regression}
\label{sec:MRL}
Now we consider a slightly different setting, where we want to learn a shared linear representation that generalizes to any target task within a set of interest.

A lot of recent papers have shown that to achieve a good generalization ability of the learned representation, the algorithm have to choose diverse source tasks \citep{tripuraneni2020theory,du2020few,xu2021representation}. They all study the performance of a given choice of source tasks, while it has been unclear whether an algorithm can adaptively select diverse tasks.

\subsection{Problem setup}
We adopt the setup in \cite{du2020few}. Let $d, k > 0$ be the dimension of input and representation, respectively ($k \ll d$). We also set $T \leq d$. Let $B^* \in \mathbb{R}^{d \times k}$ be the shared representation. Let $\beta_{1}^*, \dots, \beta_{T}^* \in \mathbb{R}^{k}$ be the linear coefficients for prediction functions. The model setup is essentially the same as the setup in Section \ref{sec:setup} except for the true parameters being $B^{*}\beta_{t}$. We call this setting structured because if one stacks the true parameters as a matrix, the  matrix has a low-rank structure.
To be specific, the output of task $t$ given by
$$
    Y_t = X_t^T B^{*}\beta_{t}^{*}   + \epsilon_t.
$$
We use the same setup for the covariate $X_t$ as in Section \ref{sec:unstructured} and we consider $\sigma_1^2 = \dots = \sigma_T^2 = \sigma^2$ for some $\sigma^2 > 0$.


\paragraph{Diversity.} Let $t_i$ be the task selected by the scheduler at step $i$. It has been well understood that to learn a representation that could generalize to any target task $t'$ with arbitrary $\beta_{t'}^*$, we will need a lower bound on the following term
\begin{equation}
    \lambda_{k}\left(\sum_{i = 1}^N \beta_{t_i}^* \beta_{t_i}^{*T}\right) \eqqcolon \lambda_{N, k}, \label{equ:condition}
\end{equation}
where $\lambda_{k}$ is the $k$-th largest eigenvalue of a matrix, i.e. the smallest eigenvalue. Basically, we hope the source tasks cover all the possible directions such that any new task could be similar to at least some of the source tasks. Equ. (\ref{equ:condition}) serves as an assumption in \cite{du2020few}. When the true $\beta_t^*$ are known, we can simply diversely pick tasks. When the $\beta_t^*$ are unknown, the trivial strategy that equally allocates samples will perform badly. For example, let $T \gg k$ and let all the $\beta_{t}, t = k + 1, \dots, T$ be identical. The trivial strategy will only cover one direction sufficiently, which ruins the generalization ability.

In this section, we will show that it is possible to adaptively schedule tasks to achieve the diversity even in the hard case discussed above. 

\subsection{Lower bounding diversity}
In this section, we introduce an OFU (optimism in face of uncertainty) algorithm that adaptively selects diverse source tasks. 

\paragraph{Two-phase estimator.} We first introduce an estimator on the unknown parameters. Assume up to step $i$, we have dataset $S_{1}^{n_{i, 1}}, \dots, S_{T}^{n_{i, T}}$ for each task $t$. We evenly split each dataset $S_{t}^{n_{i, t}}$ to two datasets $S_{i, t}^{(1)}$ and $S_{i, t}^{(2)}$, both with a sample size of $\lfloor n_{i, t} / 2 \rfloor$.
We solve the optimization problem below:
$$
    \hat B_{i} = \argmin_{B \in \mathbb{R}^{d \times k}} \min_{\beta_{t} \in \mathbb{R}^{k}, t \in [T]} \sum_{t \in [T]} \sum_{(x, y ) \in S_{i, t}^{(1)}}\|y - x^T  B \beta_{t}\|^2, 
$$
$$
    \text{ and } \hat \beta_{i, t} = \argmin_{\beta_{t} \in \mathbb{R}^{k}} \sum_{(x, y) \in S_{i, t}^{(2)}}\|y_{j} - x_{j}  B \beta_{t_j}\|^2.
$$
Note that we split the dataset such that $\hat B_i$ and $\hat \beta_{i, t}$ are independent.

\begin{algorithm}[h]
	\centering
	\caption{CL by optimistic scheduling}
	\begin{algorithmic}[1]
		\STATE \textbf{Input: } $T$ tasks and total number of observations $N$ and constant $\gamma > 0$.
		\STATE Sample $\lceil\gamma(d + \log(N/\delta))\rceil$ samples for each task.
        \FOR{$i = T\lceil\gamma(d + \log(N/\delta))\rceil + 1, \dots, N$}
        \STATE Construct confidence set $\mathcal{B}_{i, t}$ for each $t \in [T]$ according to Equ. (\ref{equ:CB}).
        \STATE Select $t_i$ according to Equ. (\ref{equ:OFU}).
        \ENDFOR
        \STATE \textbf{return:} Curriculum $(t_1, \dots, t_N)$.
	\end{algorithmic}
\label{algo:UCB}
\end{algorithm}

\paragraph{Optimistic task scheduler.} 
Our algorithm runs by keeping a confidence bound for $B^{*}\beta_t^*$ for each $t \in [T]$ and each step $i \in [N]$. Lemma \ref{lem:CB} introduces a suitable upper bound construction. Lemma \ref{lem:CB} holds under the following assumptions.

\begin{lem}
\label{lem:CB}
Let $\kappa = C_0 / C_1$. Assume  Assumption \ref{aspt:coverage} hold. There exists universal constants $\gamma > 0, \alpha > 0$ such that, at all step $i > T\lceil\gamma(d + \log(N/\delta))\rceil$, with a probability $1-\delta$, we have for all $t \in [T]$,
\begin{align*}
    &\quad\|\hat B_{i}^T \hat \beta_{i, t} - B^{*}\beta_{i, t}^*\|_2^2 
    \lesssim \frac{\alpha C_5 \sigma^2 d k \log (\kappa N \delta / T)}{C_1^2 n_{i, t}} ,
\end{align*}
where $n_{i, t}$ is the number of observations from task $t$ up to step $i$.
\end{lem}

Following the bound in Lemma \ref{lem:CB}, we construct the confidence set with width 
$$
    \mathcal{W}_{i, t} \coloneqq \frac{C_5 \sigma^2 d k \log (\kappa N \delta / T)}{C_1^2 n_{i, t}}.
$$
At each step $i$ for each task $t$, we construct a confidence set around $\hat B_i \hat \beta_{i, t}$,
\begin{equation}
    \mathcal{B}_{i, t} = \{\theta \in \mathbb{R}^{d}: \|\hat B_i \hat \beta_{i, t} - \theta\|_2^2 \leq \mathcal{W}_{i, t}\}. \numberthis \label{equ:CB}
\end{equation}

Then following the principle of optimism in face of uncertainty, we select the task $t_i$ such that 
\begin{equation}
    t_i \in \argmax_{t \in [T]} \max_{\theta \in \mathcal{B}_{i, t}} \lambda_{k} (\sum_{j = 1}^{i-1} \tilde\theta_{j} \tilde\theta_{j}^T + \theta \theta^T) \label{equ:OFU}
\end{equation}
and $\tilde\theta_i = \argmax_{\theta \in \mathcal{B}_{i, t}} \lambda_{k} (\sum_{j = 1}^{i-1} \tilde\theta_{j} \tilde\theta_{j}^T + \theta \theta^T)$. Here $\tilde\theta_i$ is our belief for task $t$ at the step $i$.

Now we are ready to present our lower bound results for diversity. Our results hold under two assumptions. The first assumption require the representation matrix $B^*$ is not degenerated. We also assume boundedness on $\beta_t^*$'s.

\begin{aspt}
\label{aspt:degen}
Assume the largest singular value of $B^*$ is smaller than $C_4$ for some $C_4 > 0$.
\end{aspt}

\begin{aspt}[Boundedness]
\label{aspt:boundedness}
We also assume that $\|\beta_{t}^*\|^2 \leq C_5$ for all $t \in [T]$.
\end{aspt}

\begin{thm}
\label{thm:div_upper}
Suppose Assumption \ref{aspt:degen} and \ref{aspt:boundedness} hold. Assume for all $\nu \in \mathbb{R}^{k}, \|\nu\|_2 = 1$, there exists some task $t$ such that $\nu^T B^* \beta^{*}_t\beta_t^{*T} B^{*T} \nu \geq \lambda$ for some $\lambda > 0$. Let $t_i, i = 1, \dots, N$ be the tasks select by Algorithm \ref{algo:UCB} for some constant $\alpha$. Then there exists some $\alpha>0$, such that with a probability at least $1-\delta$,
$$
\frac{\lambda_{N, k}}{N} \gtrsim  \frac{\lambda}{C_4k} - \sqrt{\frac{\sigma^2 C_5^2  d k T  \log (\kappa N / (T\delta))}{C_4^2 C_1^2 \lambda N}}.
$$
\end{thm}

If we are provided with the oracle, we will only have the first term above. When $N$ is sufficiently large, the second term in Theorem \ref{thm:div_upper} is negligible and we will achieve diversity asymptotically as long as $dkT \ll N$. Our proof follows the standard framework for OFU algorithms. We first show the correctness of the confidence set implied by Lemma \ref{lem:CB}. Then the key steps are to show the optimism, i.e.
$
    \lambda_{k}(\sum_{i=1}^{N} \tilde\theta_{i} \tilde\theta_{i}^{T}) = \Omega({\lambda}/{k})
$
and to bound the difference term between the belief $\lambda_{k}(\sum_{i=1}^{N} \tilde\theta_{i} \tilde\theta_{i}^{T})$ and the actual value $\lambda_{N, k}$. We provide the proof in Appendix \ref{app:thm_UCB}. 

\subsection{Upper bound results}
Though the lower bound in Theorem \ref{thm:div_upper} is already satisfying, we still want to shed some light on whether the dependency on $\sqrt{1/N}$ is avoidable by showing an upper bound result in Theorem \ref{thm:div_lower}.

\begin{thm}
\label{thm:div_lower}
For any curriculum learning algorithm, there exists $T$ tasks ($T > k$) such that for all $\nu \in \mathbb{R}^{k}, \|\nu\|_2 = 1$, there exists some $\beta_t^*$, $\|\beta_t^* \nu\| \geq 1$ and 
$$
    \mathbb{E}[\frac{\lambda_{N, k}}{N}] \lesssim \frac{\max_{t_1, \dots, t_N \in [T]}\lambda_k (\sum_{i=1}^{N}\beta_{t_i}^*\beta_{t_i}^{*T})}{N} - \sqrt{\frac{\sigma^2 T}{Nk^3}}.
$$
\end{thm}

Theorem \ref{thm:div_lower} states that the $\sqrt{1/N}$ dependency is unavoidable, while there is still a gap of $dk^4$ between the upper bound and the lower bound. Our hard case construction is inspired by the case where the naive strategy that allocates samples evenly. 
To be specific, we consider $T$ tasks such that $k$ of them are diversely specified and all the other $T - k$ tasks are identical. Naive strategies will fail by having $\lambda_{N, k} \approx \frac{1}{kT}$. We divide $T$ tasks into $\lfloor T / k \rfloor$ blocks. Then we construct similar problems. Different problems have the diverse tasks in different blocks. The difficulty of the problem becomes identifying the block with diverse tasks, which is analogous to the idea of bandit model in a general sense. From here, we follow a similar proof of stochastic bandits \citep{lattimore2020bandit}. The full proofs can be found in Appendix \ref{app:div_lower}.

\section{Analysis of Prediction Gain}
\label{sec:prediction_gain}
In this section, we give some theoretical guarantees on prediction-gain driven task scheduler under the unstructured setting discussed in Section \ref{sec:unstructured}. Wo do not consider the structured setting because it is not clear how to apply the prediction-gain driven method to multitask representation learning setting.

\paragraph{Prediction Gain and convergence rate.} 
We define prediction gain in the following way. At the step $i$, a multitask learning algorithm $\mathcal{A}$ maps any trajectory $\mathcal{\mathcal{H}}_{i} = \{x_{t_i, j}, y_{t_i, j}\}_{j = 1}^{i}$ to a parameter $\theta \in \mathbb{R}^d$ for the target task. Let the estimate at step $i$ be $\theta_i$. The prediction gain is defined as 
$$
    G(\mathcal{A}, \mathcal{H}_{i+1}) \coloneqq L_T({\theta}_{i}) - L_T({\theta}_{i+1}).
$$

At the start of the round $i$, the prediction-gain based task scheduler selects $t_i \in [T]$ such that 
$
    G(\mathcal{A}, \mathcal{H}_{i})
$ is maximized.

Note that in general, prediction gain is not observable to the algorithm before $x_{t_i, i}$ and $y_{t_i, i}$ are actually sampled. There are simple ways to estimate prediction gain, for example, from several random samples from each task.

In a linear model, the prediction gain is equivalent to convergence rate.
\begin{equation*}
L_T({\theta}_{i})-L_T({\theta}_{i+1})
    =  \|{\theta}_{i} - {\theta}^*_{T}\|_{\Sigma_T}^2 - \|{\theta}_{i+1} - {\theta}^*_{T}\|_{\Sigma_T}^2.
\end{equation*}

\cite{weinshall2020theory} discussed various benefits of curriculum learning by show that their strategy gives higher local convergence rate. It is not clear from the context that the greedy strategy that selects the highest local prediction gain gives the best total prediction gain in long run.  

\paragraph{Decomposing prediction gain.} Considering a identical covariance matrix $\Sigma_t = I$, the loss over a given parameter $\theta$ can be written as $\|\theta - \theta_T^*\|_2^2 + \sigma_T^2$.

Assume the gradient is calculated from a sample from the task $t$. According to the update of SGD, at the step $i$, we have
$$
    \theta_{i+1} - \theta_T^* = (I- \eta_i x_i^{(t)}x_i^{(t)T})(\theta_i - \theta_T^*) + \eta_i x_i^{(t)}(\epsilon_i + x_i^T\theta^{\Delta}_{t, T}),
$$
where $\theta^{\Delta}_{t, T} = \theta_t^* - \theta_T^*$.

The one-step prediction gain is 
\begin{align*}
    &\quad \|\theta_{i} - \theta_T^*\|^2 - \|\theta_{i+1} - \theta_T^*\|^2 \\
    &= \eta_t \|\theta_i - \theta_T^*\|^2_{(2-\eta_i\|x_i^{(t)}\|_2^2) x_ix_i^T} - \eta_i^2 \|x_i^{(t)}(\epsilon_i + x_i^T \theta^{\Delta}_{t, T})\|_2^2 \\
    &\quad - \eta_i(\theta_i - \theta_T^*)^T(I-\eta_i x_i^{(t)} x_i^T)x_i^{(t)}(\epsilon_i + x_i^T \theta^{\Delta}_{t, T}).
\end{align*}
The first term on the R.H.S is the absolute gain shared by all the tasks.
On expectation, the second term is
\begin{equation}
    -\mathbb{E} \eta_i^2 \|x_i (\epsilon_i - x_i^T \theta^{\Delta}_{t, T})\|_2^2 = -\mathbb{E}\eta_i^2\|x_i\|_2^2(\sigma_t^2 + \|\theta^{\Delta}_{t, T}\|_{x_ix_i^T}^2). \label{equ:term3}
\end{equation}
In expectation, the third term is 
\begin{align*}
    &\quad-\mathbb{E}\eta_{i}(\theta_{i}-{\theta}^*_{T})^{T}(I-\eta_{i} x_{i} x_{i}^{T}) x_{i}x_{i}^{T} \theta^{\Delta}_{t, T}\\
    &= -\mathbb{E}(1-\eta_i\|x_t\|_2^2)\eta_{i}(\theta_{i} - {\theta}^*_{T})^{T} x_{i}x_{i}^{T} \theta^{\Delta}_{t, T}. \numberthis \label{equ:term2}
\end{align*}

Now we discuss term (\ref{equ:term3}) and (\ref{equ:term2}), respectively. (\ref{equ:term2}) is independent of $\sigma_t^2$ and it is a dynamic effects depending on the current estimate $\theta_i$. That means (\ref{equ:term2}) is independent of the task difficulty and its constantly changes. When $(\theta_i - \theta^*_T)^T x_tx_t^T (\theta_t^* - \theta^*_T)^T < 0$, the task $t$ has a larger prediction gain. This is when the gradient descent direction is consistent in both target and the task $t$. 

For term (\ref{equ:term3}), we notice that task difficulty $\sigma_t^2$ and transfer distance $\Delta_{t, T}$ play equal importance in the prediction gain measure regardless of the number of observations. 

\paragraph{Optimality of prediction gain.} 
Let $t^*$ be the optimal task defined by
$$
    t^* = \argmin_{t}\Delta_{t, T}^2 + \frac{d\sigma_t^2}{N}.
$$
We consider an averaging SGD algorithm with a step size $\eta_i = 1/i$. In general, let $\bar\theta_N = \sum_{i = 1}^N \theta_i/N$. The following Theorem shows that the performance of the averaging SGD with an accurate prediction-gain based task scheduler matches the minimax lower bound in Theorem \ref{thm:oracle_lower}. 
\begin{thm}
\label{thm:opt_PG}
Assume $\Sigma_1 = \dots = \Sigma_T = I$. Assume $\|\theta_t^*\|_2^2 \leq C_5$ for all $t$. Given $T$ tasks with noise levels $\sigma_1^2, \dots, \sigma_T^2$ and transfer distance $\Delta_{1, T}, \dots, \Delta_{T, T}$, let $\bar \theta_N$ be the averaging SGD estimator with an accurate prediction-gain based task scheduler defined above. We have 
\begin{equation}
    G_T(\bar \theta_N) \lesssim \Delta^2_{t^*, T} + \frac{(d\sigma_{t^*}^2 + C_5)\log(N)}{N}. \label{equ:thm_opt_PG}
\end{equation}
\end{thm}

Theorem \ref{thm:opt_PG} gives an upper bound on $G_T(\bar \theta_N)$ that matches the lower bound in Theorem \ref{thm:oracle_lower}. 

\section{Discussion}
In this paper, we discussed the benefits of Curriculum Learning under two special settings: multitask linear regression and multitask representation learning. In the multitask linear regression setting, it is fundamentally hard to adaptively identify the optimal source task to transfer. In the multitask representation learning setting, a good curriculum is the curriculum that diversifies the source tasks. We show that the extra error caused by the adaptive learning is small and it is possible to achieve a near-optimal curriculum. Then we provided theoretical justification for the popular prediction-gain driven task scheduler that has been used in the empirical work.

Our results suggest some natural directions for future work. We show a lower bound (Thm. \ref{thm:div_lower}) on the diversity in the multitask representation learning setting, while leaving a gap of $d$ compared to our upper bound (Thm. \ref{thm:div_upper}). We believe this gap is because a loose construction of the hard cases that ignores the difficulty of learning the shared representation. Another direction is to show whether prediction-gain methods with no accurate gain estimation could still have performance close to lower bounds for the adaptive learning setting. 

\bibliographystyle{apalike}
\bibliography{main}

\newpage
\onecolumn

\appendix
\section{Proof of Theorem 1}

\begin{proof} 
    Our proof is inspired by the proof of \cite{kalan2020minimax}, which gives a lower bound construction for the two-tasks transfer learning problem. Our results can be seen as an extension of their constructions to multiple-source tasks setting.
    
    We define the optimal task 
    $$
        t^* = \argmin_{t}\{\bm{Q}_{t}^2 + \frac{d\sigma_t^2}{N}\}.
    $$ 
    Let $\delta^2 = (\bm{Q}_{t^*}^2 + \frac{d\sigma_{t^*}^2}{N}) / 64$. In general, we construct $T \times M$ parameters $\{\theta_{t, i}\}_{t \in [T], i \in [M]}$ with the $t$-th row corresponding to the hypothesis set of the $t$-th task. 
    
    We start by constructing the the hypothesis set of the target task and the task $t^*$. Let $\delta' = \bm{Q}_{t^*} / 16 + \delta$. By definition, we have $\delta' \leq 1.5\delta$. 
    
    Consider the set 
    $
        \Theta = \{\theta: \|\theta\|_2 \leq 2\delta'\}.
    $
    Let $\{\theta_{t^*, 1}, \dots, \theta_{t^*, M}\}$ be a $\delta'$-packing of the set in the $L_2$-norm ($\|\theta_{t^*, i} - \theta_{t^*, j}\|_2 \geq \delta'$). We can find the packing with $\log(M) \leq d\log(2)$. Since $\theta_{t^*, i}, \theta_{t^*, j} \in \Theta$, we also have $\|\theta_{t^*, i} - \theta_{t^*, j}\|_2 \leq 4\delta'$ for any $i, j \in [M]$. 
    
    Now we construct hypothesis set for the target task. For all $i \in [M]$, we choose $\theta_{T, i}$ such that $\|\theta_{T, i} - \theta_{t^*, i}\|_2 = \bm{Q}_{t^*} / 16$. So the construction for the target tasks satisfies
    $$
        \|\theta_{T, i} - \theta_{T, j}\|_2 \geq \delta' - \bm{Q}_{t^*} / 16 \geq \delta / 2 \text{\quad and \quad } \|\theta_{T, i} - \theta_{T, j}\|_2 \leq 4\delta' + \bm{Q}_{t^*} / 16 \leq 5\delta'.
    $$
    
    
    Now we discuss two cases. For any task $t$ with $\bm{Q}_{t} \geq 5\delta'$, we randomly pick a parameter in the hypothesis set of the target task which we denote by $\tilde{\theta}_t$ and we set all $\theta_{t, i} = \tilde{\theta}_t$ for all $i \in [M]$. This construction is valid since any $\|\theta_{t, i} - \theta_{T, i}\|_2 \leq 5\delta' \leq \bm{Q}_{t}$.
    
    For any task $t$ with $\bm{Q}_{t} \leq 5\delta'$, we will use the same construction as we use for $t^*$.
    
    Let $J$ be a random variable uniformly over $[M]$ representing the true hypothesis. The samples for each task $t$ is i.i.d. generated from the linear model described in Section \ref{sec:setup} with a parameter $\theta_{t, J}$. Our goal is to show that on expectation, any algorithm will perform badly as in Theorem \ref{thm:oracle_lower}.
    
    Let $E_t$ be a random sample from task $t$ given the true parameter being $\theta_{t, J}$. Similarly to (5.2) in \cite{kalan2020minimax}, using Fano's inequality, we can conclude that 
    \begin{equation}
        R_{T}^{N}(\Theta(\bm{Q})) \geq \delta^2 \left( 1 - \frac{\log(2) + \sum_{t = 1}^T n_{t} I(J; E_t)}{\log(M)}\right). \label{equ:thm1_prf_2}
    \end{equation}
    
    We proceed by giving an uniform bound on the mutual information.
    We will need the following lemma to upper bound the mutual information term.
    
    \begin{lem}[Lemma 1 in \cite{kalan2020minimax}]
    \label{lem:KL_div}
    The mutual information between $J$ and any sample $E_t$ can be upper bounded by $I(J ; E_t) \leq \frac{1}{M^{2}} \sum_{i, j} D_{K L}\left(\mathbb{P}_{\theta_{t, i}} \| \mathbb{P}_{\theta_{t, j}}\right)$, where $\mathbb{P}_{\theta_{t, i}}$ is the induced distribution by the parameter $\theta_{t, i}$. Furthermore we have
    $$
        D_{K L}\left(\mathbb{P}_{\theta_{t, i}} \| \mathbb{P}_{\theta_{t, j}}\right) = \|\Sigma_t^{1/2}(\theta_{t, i} - \theta_{t, j})\|_2^2 / (2\sigma_t^2) \leq C_0 \|\theta_{t, i} - \theta_{t, j}\|_2^2 / (2\sigma_t^2).
    $$
    \end{lem}
    \label{lem:mutual_info_bound}
    Using Lemma \ref{lem:KL_div}, we bound the mutual information of any task $t$.
    \begin{lem}
        Under the constructions introduced above, the mutual information 
        $$
            I(J; E_t) \leq \frac{512C_0}{7\sigma_{t^*}^2}\delta'^2 \text{ for all } t \in [T].
        $$ 
    \end{lem}
    \begin{proof}
    For any task in the first case discussed above ($Q_t \geq 5\delta'$), the mutual information $I(J; E_t)$ is 0. Thus the statement holds trivially.
    
    Now we discuss the second case above. By definition, we have
    \begin{equation}
        \bm{Q}^2_{t} + \frac{d\sigma_t^2}{N} \geq \bm{Q}^2_{t^*} + \frac{d\sigma_{t^*}^2}{N} = 64\delta^2. \label{equ:thm1_prf_1}
    \end{equation}
    Note that 
    $$
        \bm{Q}_{t} \leq 5\delta' = 5 (\bm{Q}_{t^*, T} / 16 + \delta) \leq 7.5\delta. 
    $$
    Plugging back into (\ref{equ:thm1_prf_1}), we have 
    $
        {d\sigma_t^2}/{N} \geq 7\delta^2,
    $
    and by definition we have ${d\sigma_{t^*}^2}/{N} \leq 64\delta^2$.
    Therefore, we have $\sigma_t^2\geq 7\sigma_{t^*}^2/(64)$.

    Since the constructions are the same for the second case, the mutual information can be uniformly bounded by
        $$
            I(J, E_t) \leq \frac{1}{M^2}\sum_{i, j} \frac{32C_0}{7\sigma_{t^*}^2} \|\theta_{t^*, i} - \theta_{t^*, j}\|_2^2 \leq \frac{512C_0}{7\sigma_{t^*}^2}\delta'^2.
        $$
    \end{proof}
    
    Finally, we follow the analysis in Section 7.4 of \cite{kalan2020minimax}. Using Lemma \ref{lem:mutual_info_bound} on Equation (\ref{equ:thm1_prf_2}), we have
    $$
        R_{T}^{N}(\Theta(\bm{Q})) \geq \delta^2 \left( 1 - \frac{\log(2) + N \frac{512C_0}{7\sigma_{t^*}^2}\delta'^2}{\log(M)}\right).
    $$
    Plugging in $\delta' = Q_{t^*}/16 + \delta$, we can conclude 
    $$
        R_{T}^{N}(\Theta(\bm{Q})) \gtrsim \bm{Q}_{t^*}^2 + \frac{d\sigma_{t^*}^2}{N}.
    $$
\end{proof}

\section{Proof of Theorem 2}
\begin{proof}

We first show the lower bound of the first term within the maximization. We construct the following problem:
we have $T-1$ tuples of parameters $\{(\theta_{1, i}, \dots, \theta_{T, i}\}_{i = 1}^{T-1}$, where $\theta_{t, i}$ corresponds to the parameters of the $t$-th task. 
Let $\{\tilde{\theta}_i\}_{i = 1}^{T-1}$ be a set of parameters that are $2\delta$-separated for some $\delta>0$. The parameters of our source and target tasks are chosen in the following manners:
\begin{enumerate}
    \item $\theta_{t, i} = \tilde{\theta}_{t}$ for all $t \in [T-1]$.
    \item $\theta_{T, i} = \tilde{\theta}_{i}$ for all $i \in [T-1]$.
\end{enumerate}
Two important properties of this construction is that 1) there is always one source task that is identical to the target task; 2) the information from source tasks can not help learn the target task.

Let $J$ follow the uniform distribution over $[T-1]$. Assume we have $n_1, \dots, n_M$ and $n_T$ be the number of observations for $T-1$ source tasks and target task from parameter $(\theta_{1, J}, \dots, \theta_{T-, J})$, respectively. 

\begin{prop}
Since $J$ is independent of $(\theta_{1, J}, \dots, \theta_{T-1, J})$, we have the mutual information $I(J; \theta_{1, J}, \dots, \theta_{T-1,J}) = 0$.
\end{prop}

Let $\psi$ be any test statistics that maps our dataset to an index. For all $\psi$, by Fano's Lemma, we can conclude that 
\begin{align*}
    \tilde{R}_T^N(\tilde{\Theta}(\bm{Q}))
    &\geq \delta^2 \frac{1}{M}\sum_{i = 1}^{M} \mathbb{P}\{\psi(S_1^{n_1}, \dots, S_M^{n_M},  S_T^{n_T}) \neq j\} \\
    &\geq \delta^2\left(1 - \frac{I(J;\psi(S_1^{n_1}, \dots, S_T^{n_T})) + \log(2))}{\log(T-1)}\right) \numberthis \label{equ:mutual}
\end{align*}

To proceed, we analyze the mutual information
\begin{align*}
    &\quad I(J;\psi(S_1^{n_1}, \dots, S_T^{n_T})) \\
    &\leq I(J;S_1^{n_1}, \dots,  S_T^{n_T})\\
    &\quad  (\text{By the independence of $S_1^{n_1}, \dots, S_{T-1}^{n_{T-1}}$ and $S_T^{n_T}$})\\
    &\leq I(J;S_1^{n_1}, \dots, S_{T-1}^{n_{T-1}}) + I(J;S_T^{n_T})\\
    &= I(J;S_T^{n_T}).
\end{align*}

Let $E$ be a random sample from the target task. We follow the analysis from \cite{kalan2020minimax}, which construct $\tilde{\theta}$ by the $2\delta$-packing of the set
$$
    \{\theta: \theta \in \mathbb{R}^{d}, \|\theta\|_2 \leq 4\delta\}.
$$
Then we can find such packing as long as $T-1 \leq d\log(2)$. 
The mutual information by the above construction gives
$$
    I(J; S_T^{n_T}) \leq n_T I(J;E) \leq n_T \frac{32\delta^2}{\sigma_T^2}.
$$

By choosing the optimal $\delta^* = \log((T-1)/2)\sigma_T^2 / (64n_T)$, we have for some $c > 0$,
$$
    \tilde{R}_T^N \gtrsim \frac{\sigma^2_T \log(T-1)}{n_T}.
$$

Note that the lower bound by Theorem \ref{thm:oracle_lower} still applies here.
In total, since $n_T \leq N$, we have 
$$
    \tilde{R}_T^N  \gtrsim \frac{{\sigma}_T^2 \log(T)}{N} + \min_t\frac{d{\sigma}_t^2}{N}.
$$
The above analysis only works when $Q_{sub}^2 \geq \delta^* =  \log((T-1)/2)\sigma_T^2 / (64n_T)$. Otherwise, one will at least suffer $Q_{sub}^2$ plus the learning difficulty term $\min_t\frac{d{\sigma}_t^2}{N}$.
\end{proof}

\section{Proof of Theorem \ref{thm:adp_upp}}
\label{app:thm3}

The proof of Theorem \ref{thm:adp_upp} is similar to many proofs of generalization bound. We let the empirical loss on the target task be
$$
    \hat L_T (\theta) = \frac{2}{N}\sum_{i=1}^{N / 2}\left(Y_{T, i}-X_{T, i}^{T} \hat{\theta}_{t}\right)^{2}.
$$
Write $Y_{T, i} = X_{T, i}^T \theta_T^* + \epsilon_{T, i}$. Let $\hat\Sigma_T = \sum_{i = 1}^{n_T} X_{T, i}X_{T, i}^T$ be the sample covariance matrix.
We start with
\begin{align*}
    &\quad L_T(\hat \theta_{t^*}) - \min_{t \in [T - 1]} L_T(\theta_t^*)\\
    &\leq L_T(\hat \theta_{t^*}) - \min_{t \in [T - 1]} L_T(\hat \theta_t^*) + \max_{t \in [T - 1]} (L_T(\hat \theta_t^*) - L_T(\theta_t^*)) \\
    &\leq \hat L_T(\hat \theta_{t^*}) - \min_{t \in [T - 1]} \hat L_T(\hat \theta_t^*) + \max_{t \in [T - 1]} (L_T(\hat \theta_t^*) - L_T(\theta_t^*)) + 2 \max_{t \in [T - 1]} |L_T(\hat \theta_{t}) - \hat L_T(\hat \theta_{t})|\\
    &=\max_{t \in [T - 1]} |L_T(\hat \theta_t) - L_T(\theta_t^*)| + 2 \max_{t \in [T - 1]} |L_T(\hat \theta_{t}) - \hat L_T(\hat \theta_{t})|,
\end{align*}
where the last equality is based on the definition of $t^*$.

Now we bound the two difference term separately. Let $n_T = N / 2$.
\begin{lem}
With a probability at least $1-\delta$, we have 
$$
    |L_T(\hat \theta_t) - \hat L_T(\hat \theta_t)| \lesssim \frac{C_0C_1C_2p\log(T/\delta)\sigma_T^2}{n_T} + \sqrt{\frac{p+\log(\delta)}{n_T}} \text{ for all } t \in [T-1].
$$
\end{lem}
\begin{proof}

We make the following decomposition. 
\begin{align*}
    &\quad |L_T(\hat \theta_{t}) - \hat L_T(\hat \theta_{t})| \\
    &= |\|\hat \theta_t - \theta^*_{T}\|_{\Sigma_T}^2 + \sigma_T^2 - \frac{1}{n_T} \sum_{i = 1}^{n_T}[X_{T, i}^T(\hat \theta_t - \theta_T^*) - \epsilon_{T, i}]^2| \\ 
    &\leq \|\hat \theta_t - \theta^*_{T}\|_{\Sigma_T - \hat \Sigma_T}^2 + |\sigma_T^2 - \frac{1}{n_T} \sum_{i = 1}^{n_T} \epsilon^2_{T, i}| + \frac{1}{n_T}| \sum_{i = 1}^{n_T}X_{T, i}^T(\hat \theta_t - \theta_T^*)\epsilon_{T, i}|.
\end{align*}

Now we bound the three terms above separately. The second term is the concentration for $\chi^2(n_T)$ distribution. We have with a probability at least $1-\delta/(3T)$, 
$
    |\sigma_T^2 - \frac{1}{n_T} \sum_{i = 1}^{n_T} \epsilon^2_{T, i}| \lesssim \sigma_T^2(\sqrt{\log(3T/\delta)/n_T} + \log(3T/\delta) / n_T).
$

To proceed, we consider the concentration of sample covariance matrix.
\begin{lem}[\cite{tropp2015introduction}]
    \label{lem:cov_concentration}
    Let $x_1, \dots, x_n \in \mathbb{R}^p$ be an i.i.d sequence of $\sigma$ sub-Gaussian random vectors such that $\operatorname{Var}(x_1) = \Sigma$ and $\hat \Sigma \coloneqq \frac{1}{n} \sum_{i = 1}^n x_ix_i^T$ be the empirical covariance matrix. Then  with a probability at least $1-\delta$, 
    $$
        \|\hat \Sigma - \Sigma\|_{op} \lesssim \sqrt{\frac{d+\log (2 / \delta)}{n}} + \frac{d+\log (2 / \delta)}{n},
    $$
    where $\|A\|_{op} = \max_{x \in \mathbb{S}^{d-1}}|x^T A x|$.
\end{lem}

Using the boundedness of both $\hat \theta_t$ and $\theta_T^*$, $\|\hat \theta_t - \theta^*_T\|_2 \lesssim C_2$.  Then applying Lemma \ref{lem:cov_concentration}, we have with a probability at least $1-\delta/(3T)$,
$$
    \|\hat \theta_t - \theta^*_T\|_{\Sigma_T - \hat \Sigma_T}^2 \lesssim C_1 (\sqrt{\frac{d+\log (2 / \delta)}{n_T}} + \frac{d+\log (2 / \delta)}{n_T}).
$$

For the third term, we show that with a probability at least $1-\delta/6$, $\|\hat \theta_t - \theta_T^*\|_{X_{T, i} X_{T, i}^T} \lesssim C_1C_2\log(n_T/\delta)$ for all $i = 1, \dots, n_T$. Then we apply the concentration inequality on the average of Gaussian noise $\frac{1}{n_T} \sum_{i = 1}^{n_T} \epsilon_{T, i}$, we have with a probability at least $1-\delta/3$, we have for all $t \in [T-1]$
$$
    \frac{1}{n_T} |\sum_{i = 1}^{n_T}X_{T, i}^T(\hat \theta_t - \theta_T^*)\epsilon_{T, i}| \lesssim \frac{C_1C_2 \sigma^2_T \log(n_T / \delta)}{n_T}.
$$

\end{proof}

\begin{lem}
\label{lem:6}
With a probability at least $1-\delta$, we have
$$
    |L_T(\hat \theta_{t}) - L_T(\theta_{t}^*)| \leq \frac{TC_0d\sigma_t^2 \log(T/\delta)}{C_1N}.
$$
\begin{proof}
Lemma \ref{lem:6} can be derived from the high probability bound of $\|\hat \theta_t - \theta_T^*\|_2$ and the Assumption \ref{aspt:coverage}.
\end{proof}
\end{lem}

\section{Proof of Proposition \ref{prop:1}}
\begin{proof}
The target task is basically minimizing the empirical loss over $T-1$ estimators. We first apply the standard generalization bound with Radermacher complexity
\begin{align*}
    L_T(\hat f_{t^*}) \leq \min_{t \in [T-1]} L_T(\hat f_{t^*}) + \sqrt{\frac{2\log(T-1)}{N}} + c\sqrt{\frac{2\log(1/\delta)}{N}},
\end{align*}
where $c$ is a universal constant.


To proceed, we bound 
$\min_{t \in [T-1]} L_T(\hat f_{t^*})$. By the assumption, we have
$L^* = \min_t L_T(f_t^*).$ Let the task that realizes the minimization be $t'$. Using Assumption \ref{aspt:gen}, we have 
\begin{align*}
    \min_{t \in [T-1]} L_T(\hat f_{t}) \leq L_T(\hat f_{t'}) = \mathbb{E}_{(X_T, Y_T)} l(\hat f_{t'}, Y_T) \leq L^* + L_1\mathbb{E}_{X_T} \|\hat f_{t'} - f_{t'}^*\|^2 \leq L^* + \frac{L_1}{L_2}(L_{t'}(\hat f_{t'}) - L_{t'}^*).
\end{align*}
We can apply the generalization bound on $L_{t'}(\hat f_{t'}) - L_{t'}^*$, which gives us the result.

\end{proof}

\section{Proof of Theorem \ref{thm:div_upper}}
\label{app:thm_UCB}

\subsection{Proof of Lemma \ref{lem:CB}} We will borrow some techniques from \cite{du2020few} for the proof of Theorem \ref{thm:div_upper}. We start with the proof of Lemma \ref{lem:CB}, which provides a valid confidence set for the unknown parameters $B^*\beta_{t}^*$.
First, we let $X_{i, t}^{(1)}$ be the covariance matrix of the first split $S_{i, t}^{(1)}$.

\begin{claim}[Covariance concentration on the first split.]
\label{clm:1}
For $\delta \in (0, 1)$, there exists a constant $\gamma_1 > 0$ such that with a probability at least $1-\delta/10$, we have 
$$
    0.9 \Sigma_{t} \prec \frac{2}{n_{i, t}} X_{i, t}^{(1) T} X_{i, t}^{(1)} \prec 1.1 \Sigma_t \text{\quad for all $i \in\{i' \in [N]: n_{i', t}\geq \gamma_1(d + \log(N/\delta))\}$}.
$$
\end{claim}

\begin{claim}[Covariance concentration on the second split.]
\label{clm:2}
For $\delta \in (0, 1)$, there exists some $\gamma_2 > 0$ such that for any given $B \in \mathbb{R}^{d \times 2k}$ that is independent of $X_{i, t}^{(2)}$, with a probability at least $1-\delta/10$, we have 
$$
    0.9 B^T \Sigma_{t} B \prec \frac{2}{n_{i, t}} B^T X_{i, t}^{(2) T} X_{i, t}^{(2)} B \prec 1.1 B^T \Sigma_t B \quad \text{for all $i \in \{i'\in[N]: n_{i', t} \geq \gamma_2 (d + \log(N/\delta))$}.
$$
\end{claim}
Let $\gamma = \max\{\gamma_1, \gamma_2\}$. Recall that $\kappa = C_0 / C_1$. Then the good events in Claim \ref{clm:1} and \ref{clm:2} hold for all $i \geq \lceil\gamma(d + \log(N/\delta))\rceil$. We first apply the Claim A.3 in \cite{du2020few}, which guarantees the loss on the source training data. We rephrase it here as Lemma \ref{lem:claimA3}. Note that the only difference is that we require the good events hold for all $i: n_{i, t}$'s are sufficiently large.
\begin{lem}[Claim A3 in \cite{du2020few}]
\label{lem:claimA3}
With a probability at least $1-\delta/5$, we have
\begin{equation}
    \sum_{t = 1}^{T} \|X_{i, t}^{(1)}(\hat B_{i} \hat \beta_{i, t} - B^*\beta_{t}^*)\|_2^2 \lesssim \sigma^{2}\left(k T+ d k \log(\kappa i/T)+\log (N / \delta)\right) \text{\quad for all $i >  \lceil\gamma(d + \log(N/\delta))\rceil$}.
    \label{equ:claimA3}
\end{equation}
\end{lem}

Note that $X_{i, t}^{(1)} \hat B_{i} \hat \beta_{i, t} = P_{X_{i, t}^{(1)}\hat B_{i}} Y_t = P_{X_{i, t}^{(1)}\hat B_{i}} (X_{i, t}^{(1)} B^* \beta_{t}^* + z_t)$. To proceed, for any fixed $t' \in [T]$, we have
\begin{align*}
    &\quad \sigma^{2}\left(k T+ d k \log(\kappa i/T)+\log (N / \delta)\right)\\
    &\gtrsim \sum_{t = 1}^{T} \|X_{i, t}^{(1)}(\hat B_{i} \hat \beta_{i, t} - B^*\beta_{t}^*)\|_2^2 \\
    &= \sum_{t = 1}^T \|P_{X_{t}\hat B_{i}}(X_{i, t}^{(1)}B^*\beta_t^* + z_t) \|_2^2 \\
    &\geq \sum_{t = 1}^T \|P_{X_{t}\hat B_{i}}X_{i, t}^{(1)}B^*\beta_t^* \|_2^2 \\
    &\geq 0.9 \sum_{t = 1}^T \frac{n_{i, t}}{2} \|P_{\Sigma_{t}^{1/2}\hat B_{i}}\Sigma_{t}B^*\beta_t^* \|_2^2 \quad  \text{(\text{Using Claim \ref{clm:1}})} \\ 
    &\geq 0.45 C_1 \sum_{t = 1}^T n_{i, t} \|P_{\Sigma_{t'}^{1/2}\hat B_{i}}\Sigma_{t'}B^*\beta_t^* \|_2^2 \\
    &= 0.45 C_1 \sum_{j = 1}^i \|P_{\Sigma_{t'}^{1/2}\hat B_{i}}\Sigma_{t'}B^*\beta_{t_j}^* \|_2^2
\end{align*}

Then we have 
\begin{align*}
    &\quad \|\hat B_{i} \hat \beta_{t'} - B^* \beta^*_{t'} \|_2^2 \\
    &\leq \frac{1}{C_1}\|\Sigma_{t'}^2(\hat B_{i} \hat \beta_{t'} - B^* \beta^*_{t'})\|_2^2 \\
    &\leq \frac{1}{0.9C_1n_{i, t'}}\|X_{i,t'}^{(2)}(\hat B_{i} \hat \beta_{t'} - B^* \beta^*_{t'})\|_2^2 \quad \text{(Using Claim \ref{clm:2})}\\ 
    &= \frac{1}{0.9C_1n_{i, t'}}\left(\|P_{X_{i,t'}^{(2)}\hat B_{i}}X_{i,t'}^{(2)}B^*\beta_{t'}^*\|_2^2 + \|P_{X_{i,t'}^{(2)}\hat B} z_{t'}\|_2^2\right)
\end{align*}
For the second term above, 
$$
    \frac{1}{\sigma^2}\|P_{X_{i,t'}^{(2)}\hat B} z_{t'}\|_2^2 \sim \chi^2(k).
$$
Thus with a probability at least $1-\delta$, $\|P_{X_{i,t'}^{(2)}\hat B} z_{t'}\|_2^2 \lesssim k + \log(NT/\delta)$ for all $t' \in [T]$ and $i > T\lceil\gamma(d + \log(N/\delta))\rceil$. Therefore, we obtain the bound: for all $i > T\lceil\gamma(d + \log(N/\delta))\rceil$ and all $t' \in [T]$, it holds that
\begin{align*}
    \|\hat B_{i} \hat \beta_{t'} - B^* \beta^*_{t'} \|_2^2 
    &\lesssim \frac{\sigma^2}{C_1}\left(\frac{k T+d k \log (\kappa i / T)+\log (N / \delta)}{C_1 n_{i, t'}/C_5}  + \frac{k + \log(NT/\delta)}{n_{i, t'}}\right)\\
    &\lesssim
     \frac{C_5 \sigma^2 d k \log (\kappa N \delta / T)}{C_1^2 n_{i, t'}}.
\end{align*}

\subsection{Proof of the full theorem} Now we prove the full theorem.
By Assumption \ref{aspt:degen}, we convert our target $\lambda_d(\sum_{i = 1}^N \beta_{t_i}^*\beta_{t_i}^{*T})$ to $\lambda_d(\sum_{i = 1}^N B^{*}\beta_{t_i}^*\beta_{t_i}^{*T}B^{*T})$:
\begin{align*}
    &\quad  \frac{1}{N}\lambda_d(\sum_{i = 1}^N \beta_{t_i}^*\beta_{t_i}^{*T}) \geq\frac{1}{NC_4}\lambda_d(\sum_{i = 1}^N B^{*}\beta_{t_i}^*\beta_{t_i}^{*T}B^{*T}).
\end{align*}
Then we follow the standard decomposition framework of UCB analysis:
\begin{equation}
    \frac{1}{N}\lambda_d(\sum_{i = 1}^N B^{*}\beta_{t_i}^*\beta_{t_i}^{*T}B^{*T}) = \frac{1}{N}\left(\lambda_{k}(\sum_{i = 1}^{N} \tilde{\theta}_i \tilde{\theta}_i^T) + \lambda_{k}(\sum_{i = 1}^{N} B^{*T}\beta^*_{t_i} \beta_{t_i}^{*T}B^{*}) - \lambda_{k}(\sum_{i = 1}^{N} \tilde{\theta}_i \tilde{\theta}_i^T)\right). \label{equ:thm4_decomp}
\end{equation}
Our proof proceeds by first showing 
$$
    \frac{1}{N}\lambda_{k}(\sum_{i = 1}^{N} \tilde{\theta}_i \tilde{\theta}_i^T) \geq \lambda/d,
$$
which is usually interpreted as optimism.
Then we bound the difference term 
$$
    \frac{1}{N}\left(\lambda_{k}(\sum_{i = 1}^{N} B^{*}\beta^*_{t_i} \beta_{t_i}^{*T}B^{*T}) - \lambda_{k}(\sum_{i = 1}^{N} \tilde{\theta}_i \tilde{\theta}_i^T)\right),
$$
which is expected to vanish when $N$ becomes large.

\paragraph{Proof of optimism.}
We apply Lemma \ref{lem:CB} and have $\tilde \theta_i \in \mathcal{B}_{i, t}^{\alpha}$. Since $B^*\beta^*_{t} \in \mathcal{B}_{i, t}^{\alpha}$ for all $t \in [T], i \in [N]$, it is easy to show that the greedy selection over $\mathcal{B}_{i, t}^{\alpha}$ will lead to
$$
    \lambda_{k}(\sum_{i=1}^{N} \tilde{\theta}_{i} \tilde{\theta}_{i}^{T}) \geq \lambda(\frac{N}{k}-1).
$$

We prove by induction. Assume at any step $n$, we have for all $\|\nu\|_2 = 1$, 
$$
    \nu^T\sum_{i=1}^{n} \tilde{\theta}_{i} \tilde{\theta}_{i}^{T}\nu \geq \lambda(i/k - 1).
$$
We will show that at the step $n + k$, we will at least have 
$$
    \nu^T\sum_{i=1}^{n} \tilde{\theta}_{i} \tilde{\theta}_{i}^{T}\nu \geq \lambda(i/k).
$$
The proof is simple, if there exists a $\nu$ such that the above inequality fails, we will select a task that brings it to $\lambda (i/k)$. This process can be done at most $k$ times.

\paragraph{Upper bounding the differences.}
We first write the difference of eigenvalues in terms of the difference of the matrices. We will use a trick here. 
\begin{align*}
&\quad\lambda_{k}(\sum_{i = 1}^{N} B^{*}\beta^*_{t_i} \beta_{t_i}^{*T}B^{*T}) - \lambda_{k}(\sum_{i = 1}^{N} \tilde{\theta}_i \tilde{\theta}_i^T)\\
&= \min_{\|\nu\|_2 = 1} \nu^T \sum_{i = 1}^{N} B^{*}\beta^*_{t_i} \beta_{t_i}^{*T}B^{*T} \nu - \min_{\|\nu\|_2 = 1} \nu^T \sum_{i = 1}^{N} \tilde{\theta}_i \tilde{\theta}_i^T \nu\\
&\geq \min_{\|\nu\|_2 = 1} \nu^T \sum_{i = 1}^{N} B^{*}\beta^*_{t_i} \beta_{t_i}^{*T}B^{*T} \nu - \min_{\|\nu\|_2 = 1} \nu^T \sum_{i = 1}^{N} \tilde{\theta}_i \tilde{\theta}_i^T \nu\\
&\geq \min_{\|\nu\|_2 = 1} \left( \nu^T \sum_{i = 1}^{N} (B^{*}\beta^*_{t_i} \beta_{t_i}^{*T}B^{*T} - \tilde{\theta}_i \tilde{\theta}_i^T) \nu \right)\\
&\geq \sum_{i = 1}^{N} \min_{\|\nu\|_2 = 1}  \nu^T  (B^{*}\beta^*_{t_i} \beta_{t_i}^{*T}B^{*T} - \tilde{\theta}_i \tilde{\theta}_i^T) \nu \\
&\geq - \sum_{i=1}^N\| B^{*}\beta^*_{t_i}  - \tilde{\theta}_i \|_2 (\|B^{*}\beta^*_{t_i}\|_2 + \|\tilde{\theta}_i\|_2)\\
&\geq - 2C_5\sum_{i=1}^N\| B^{*}\beta^*_{t_i}  - \tilde{\theta}_i\|_2.
\end{align*}

Applying Lemma \ref{lem:CB} and by the construction of the confidence set $\mathcal{B}_{i, t}^{\alpha}$, we have 
$$
    \| B^{*}\beta^*_{t_i}  - \tilde{\theta}_i\|_2 \lesssim 
    \sqrt{\frac{C_5 \sigma^2 d k \log (\kappa N \delta / T)}{C_1^2 n_{i, t}}}.
$$
Thus,
$$
    \lambda_{k}(\sum_{i = 1}^{N} B^{*}\beta^*_{t_i} \beta_{t_i}^{*T}B^{*T}) - \lambda_{k}(\sum_{i = 1}^{N} \tilde{\theta}_i \tilde{\theta}_i^T) \gtrsim
    -\sum_i\sqrt{\frac{C_5^2 \sigma^2 d k \log (\kappa N \delta / T)}{C_1^2 n_{i, t}}} \gtrsim \sum_i\sqrt{\frac{C_5^2 \sigma^2 d k TN \log (\kappa N \delta / T)}{C_1^2}}.
$$

Plugging this back to the decomposition term (\ref{equ:thm4_decomp}) we arrive the final bound.

\section{Proof of Theorem \ref{thm:div_lower}}
\label{app:div_lower}


Assume we have $T$ tasks in total. We pick a set of orthogonal vectors $\{\beta_1, \dots, \beta_k\} \in \mathbb{R}^{k}$ with $\|\beta_i\|_2^2 = \lambda$. We first construct a simple instance in the following way: the first $k$ tasks are diverse such that $(\beta_{i}^*, \dots, \beta_{k}^*) = (\beta_1, \dots, \beta_k)$. Then all the other tasks share the same parameter $\beta_{1}$. We denote the instance by $v$. This construction is hard for naive task scheduler that evenly allocates samples to all the tasks since the direction for $\beta_1^*$ will be over-exploited. 

We evenly divide $T$ tasks into $M = \lfloor T / k \rfloor$ blocks. Let $T_m$ be the total number of visits in the $m$-th block. For any task scheduler, there exists $m' \in [M]$ such that $\mathbb{E}[T_m] \leq \frac{N}{M}$ by pigeonhole theorem.

Then we construct another instance denoted by $v'$ such that $v$ is the same as $v'$ except for that the $m$-th block has the parameters $(2\beta_1, \dots, 2\beta_k)$ for the $k$ tasks in the block.

Let $P_{v}$ and $P_{v'}$ be the probability measure on the linear regression model with true parameter defined in Section \ref{sec:setup} for $v$ and $v'$.

Define $\Delta_{N, k}(\mathcal{T}, v)$ be the expected difference using task scheduler $\mathcal{T}$ on instance $v$, i.e.
$$
    \Delta_{N, k}(\mathcal{T}, v) \coloneqq \max_{t_1, \dots, t_N} \lambda_{k}(\sum_{i = 1}^N \beta_{t_i}^* \beta_{t_i}^{*T}) - \lambda_{N, k}.
$$

Thus, applying Bretagnolle–Huber inequality (Theorem 14.2 \citep{lattimore2020bandit}) we have
$$
    \Delta_{N, k}(\mathcal{T}, v) + \Delta_{N, k}(\mathcal{T}, v') \geq \frac{N\lambda}{2k}(P_{v}(T_1 \leq N / M) + P_{v'}(T_1 > N / M)) \geq \frac{N\lambda}{2k}\exp(-D(P_{v}, P_{v'})).
$$
where $D(P, Q)$ is the relative entropy between distributions $P$ and $Q$. 

Then we apply Lemma 15.1 \citep{lattimore2020bandit}, which we rephrase here. 
\begin{lem}
Let $P_t$ and $P_t'$ be the probability measure of the $t$-th task using true parameters from $v$ and $v'$, respectively. We also let $\bar T_t$ be the number of observations on the $t$-th task. Then we have
$$
    D(P_{v}, P_{v'}) = \sum_{t = 1}^T \mathbb{E}_{v}[\bar T_t] D(P_t, P_t') \leq \mathbb{E}_{v}[T_m] \max_{t = m(k-1) + 1, \dots, mk}{D(P_t, P_{t}')}.
$$
\end{lem}

Now since $D(P_t, P_{t}') = {\|\beta_t - \beta_t^*\|^2}/(2\sigma^2) = \lambda^2/(2\sigma^2)$, we have
$$
    \Delta_{N, k}(\mathcal{T}, v) + \Delta_{N, k}(\mathcal{T}, v') \geq \frac{N\lambda}{2k}\exp(\frac{N\lambda^2}{2M\sigma^2}).
$$

Choosing $\lambda = \frac{2M\sigma^2}{N}$, we have
$$
    \Delta_{N, k}(\mathcal{T}, v) + \Delta_{N, k}(\mathcal{T}, v') \geq \sigma\sqrt{NM}/k = \sigma\sqrt{NT/k^3}.
$$

\section{Proof of Theorem \ref{thm:opt_PG}}

\begin{proof}
We follow the standard procedure of the convergence analysis of SGD. Let $t_i$ be the task that the task scheduler chooses at the step $i$. Let $v_i^{(t)}$ be the virtual gradient calculated at the step $i$ if task $t$ is scheduled, i.e.
$$
    v_i^{(t)} = x_i^{(t)T} (\theta_{t}^* -  \theta_i) x_i^{(t)} + \epsilon_i^{(t)} x_i^{(t)},
$$
where $\epsilon_i^{(t)}$ and $x_i^{(t)}$ is the random noise and the input sampled at the step $i$ from task $t$.
To start with, let $\theta_{i+1}^{(t)}$ be the virtual next step if task $t$ is scheduled. We have
$$
    \theta_{i + 1}^{(t)} - \theta_T^* = \theta_i - \theta_T^* - \eta_i v_i^{(t)}.
$$
By algebra, we derive
\begin{align*}
    &\quad \|\theta_{i + 1}^{(t)} - \theta_T^*\|^2 - \|\theta_i - \theta_T^*\|^2 \\
    &= -2\eta_i (\theta_i - \theta_T^*)^T x_i^{(t)}x_i^{(t)T}(\theta_i - \theta_{T}^*) + \\
    &\quad 2\eta_i (\theta_i - \theta_T^*)^T x_i^{(t)}x_i^{(t)T}(\theta_{t}^* - \theta_{T}^*) - \\
    &\quad 2\eta_i  x_i^{(t)T}(\theta_i - \theta_T^*) \epsilon_{i} + \eta_i^2 \|v_i^{(t)}\|^2.
\end{align*}
Taking expectations over $x_i^{(t)}$ and arrange the equation, we have
\begin{align*}
    \|\theta_i - \theta_T^*\|^2
    =  \frac{\|\theta_i - \theta_T^*\|^2 - \|\theta_{i + 1}^{(t)} - \theta_T^*\|^2}{2\eta_i} 
      +
     (\theta_i - \theta_T^*)^T (\theta_{t}^* - \theta_{T}^*) + \frac{\eta_i}{2} \|v_i^{(t)}\|^2. \numberthis
      \label{equ:thm6_1}
\end{align*}
Note that 
$$
    L_T(\theta_i) - \sigma_T^2 = \|\theta_i - \theta_T^*\|^2.
$$ 
Plugging this into Equ. (\ref{equ:thm6_1}), we have 
\begin{align*}
     L_T(\theta_i) - \sigma_T^2 
    =  \frac{\|\theta_i - \theta_T^*\| - \|\theta_{i + 1}^{(t)} - \theta_T^*\|^2}{2\eta_i} + 
      (\theta_i - \theta_T^*)^T (\theta_{t}^* - \theta_{T}^*) + \frac{\eta_i}{2} {\|v_i^{(t)}\|^2} .
\end{align*}
Since this holds for any $t$, we let $t = t^*$ and note that by the definition of the task scheduler with accurate prediction gain estimate, 
$$
    \mathbb{E}\|\theta_i - \theta_T^*\|^2 \leq \mathbb{E}\|\theta_i^{(t)} - \theta_T^*\|^2.
$$
We have 
\begin{align*}
    &\quad \mathbb{E}[L_T(\theta_i)] - \sigma_T^2 \\
    &\leq \mathbb{E} \frac{\|\theta_i^{(t^*)} - \theta_T^*\|^2 - \|\theta_{i + 1}^{(t^*)} - \theta_T^*\|^2}{2\eta_i} + 
    \mathbb{E} (\theta_i - \theta_T^*)^T(\theta_{t^*}^* - \theta_{T}^*) + \frac{\eta_i}{2} \mathbb{E}{\|v_i^{(t^*)}\|^2}.
\end{align*}
Note that taking $\eta_i = 1/i$, the first term on the right hand side collapses to $-N\E\|\theta_{N+1}^{(t)} - \theta_T^*\|^2 \leq 0$.

To proceed, we have 
\begin{align*}
    &\quad \sum_{i = 1}^N \mathbb{E}(\theta_i - \theta_T^*)^T (\theta_i - \theta_{T}^*)\\ &\leq \sum_{i = 1}^N \mathbb{E}(\theta_i - \theta_T^*)^T (\theta_{t^*}^* - \theta_{T}^*) + \sum_i\frac{\eta_i}{2} \mathbb{E}{\|v_i^{(t^*)}\|^2} \\
    &\leq \sum_{i = 1}^N \mathbb{E}(\theta_i - \theta_T^*)^T (\theta_{t^*}^* - \theta_{T}^*) + \log(N) (\sigma_{t^*}^2 d + C_5)\\
    &\leq \sqrt{\sum_{i = 1}^N \mathbb{E}\|\theta_i - \theta_T^*\|^2} \sqrt{\sum_{i = 1}^N \|\theta_{t^*}^* - \theta_{T}^*\|^2} + \log(N) (\sigma_{t^*}^2d + C_5).
\end{align*}
By solving the inequality, we have 
\begin{align*}
    \sum_{i = 1}^N \mathbb{E}\|\theta_i - \theta_T^*)^T\|^2 &\leq \sum_{i = 1}^N \mathbb{E}\|\theta_{t^*}^* - \theta_{T}^*\|^2 + \log(N)(\sigma_{t^*}^2d + C_5)\\
    &\leq \sum_{i = 1}^N \Delta_{t^*, T}^2 + \log(N)(\sigma_{t^*}^2d + C_5)
\end{align*}
Divided by $N$ on both side, we reach Theorem \ref{thm:opt_PG}.
\end{proof}

\section{Simple Empirical Verification}
We consider five linear regression tasks with $\sigma^2 = 0.05, 0.1, 0.5, 1, 2$, respectively. The dimension $d$ is set to be 3. We randomly sample each dimension of the true coefficients from $N(0, 0.1)$. The last task is considered as the target task. The distance $\Delta_{t, 5}$'s are calculated. We allow total observations $N = 1000$. 

We designed the below optimistic simulation. The base optimization algorithm runs SGD with fixed learning rate ($\eta_i = 1/(di), i = 1, \dots, 1000$) on the samples provided by task scheduler. We sample 1000 observations from each task to have a fixed dataset. Any task scheduler decides a task to sample and the next unsampled observation in its dataset will be fed to SGD.

We compare two task schedulers: 1) prediction-gain-driven scheduler with an accurate prediction (by accurate prediction, we do one virtual step of gradient descent with a sample from each task and evaluate the actual prediction gain on the true parameters); 2) a fixed scheduler that always select the optimal task indicated by Theorem \ref{thm:oracle_lower}. 

The average MSE is given by 0.002069853 and 0.003441738 for 1) and 2) respectively.

\end{document}